\documentclass[letterpaper, 10 pt, conference]{ieeeconf}  

\IEEEoverridecommandlockouts                              

\usepackage{graphics} 
\usepackage{epsfig} 
\usepackage{mathptmx} 
\usepackage{times} 
\usepackage{amsmath} 
\usepackage{amssymb}  
\usepackage{cite}

\usepackage{algorithm}
\usepackage{algpseudocode}
\usepackage{gensymb}
\usepackage{todonotes}
\usepackage{url}      
\title{\LARGE \bf
Tendon-Driven Continuum Robot with Modular Stiffness and In-Situ Self Pose Estimation
}

\author{Guo Ning (Andrew) Sue$^{*}$, Zheng Cao$^{*}$, Junzhe Hu$^{*}$,  Xiangyun Bu, David Quinn, \\ Tiancheng Wu, Zackory Erickson, Carmel Majidi
\thanks{*Equal Contribution}
\thanks{All authors are from Carnegie Mellon University \tt\footnotesize{\{gsue, 	
zhengcao, junzhehu, xiangyun, djquinn, tianche3, zachory, cmajidi\}@andrew.cmu.edu}}}

\begin{document}

\maketitle
\thispagestyle{empty}
\pagestyle{empty}

\begin{abstract}

Continuum robots enable smooth shape morphing and safe interaction in confined environments.  However, most existing systems are task-specific and depend on external sensing infrastructure, limiting their adaptability and real-world deployment. This paper presents a self-contained modular continuum robotic platform that combines mechanical reconfigurability with onboard pose estimation. The robot is constructed from interchangeable continuum joints with analytically precomputed stiffness, allowing rapid assembly and direct programming of the robot shape. Proprioceptive sensing is achieved using magnetic sensors and a modular learning-based framework, where a single model is trained per joint and reused across configurations. The system is experimentally validated in real world, demonstrating self-sensing capabilities and adaptation without external tracking.

\end{abstract}

\section{INTRODUCTION}
Continuum robots have gained attention for their compliance and adaptability compared to rigid-link robots, enabling safe interaction and smooth continuous deformation inspired by biological structures such as octopus arms and elephant trunks \cite{russo2023continuum}. These traits support dexterous manipulation and navigation in confined or cluttered spaces, making them suitable for search \cite{yamauchi2022development}, minimally invasive surgery \cite{burgner2015continuum}, inspection \cite{wang2021design}, and human–robot interaction \cite{abah2021multi}.

\begin{figure}[!t]
\centering
    \includegraphics[totalheight=7cm]{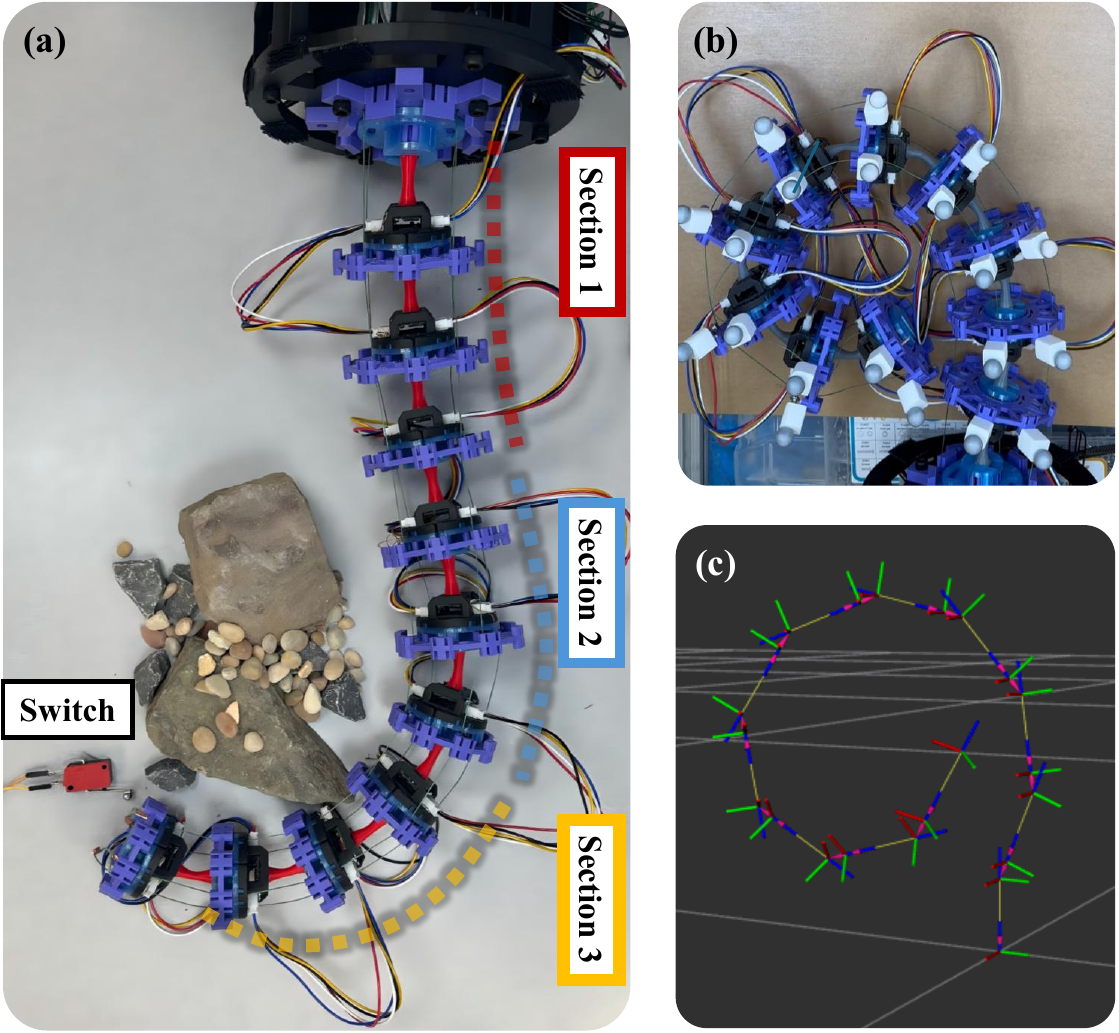}
    \caption{(a) Example of our modular, variable stiffness, continuum robot platform.  In this example, a 10 segment robot with  joints of 
"high" (Section 1), "medium" (Section 2) and "low" (Section 3) stiffness is used to produce a shape that can reach around an obstacle and contact a switch. (b) Example of a large-deformation configuration of the robot under rotational motion; (c) Corresponding pose estimation result for the configuration shown in (b).}
    \label{fig:overall}
\end{figure}

Most existing continuum robots are still task-specific and built in an ad hoc way, with fixed geometry, actuation, and mechanical properties that are hard to reconfigure. As a result, adapting them to new tasks often requires major redesign and refabrication, limiting scalability, and broader adoption. This lack of modularity stands in contrast to the versatility that continuum robots are theoretically capable of providing.

Other major barriers to the practical deployment of continuum robots are sensoring and state estimation. Unlike rigid robots with a small set of joint variables, continuum robots effectively have infinite degrees of freedom, making shape and pose estimation difficult. Many approaches therefore rely on external infrastructure (e.g., motion capture, cameras, or tracking rigs) \cite{sincak2024sensing, lu2023image}, which works in the lab but increases cost and complexity and limits use outside controlled environments.

To address these limitations, we present a self-contained modular continuum robotic platform that combines mechanical reconfigurability with embedded proprioceptive sensing. The proposed system is designed around modular continuum joints that can be rapidly assembled, replaced, or reconfigured to meet different task requirements without sacrificing the fundamental advantages of continuum robots, namely their smooth shape morphing and compliance. The mechanical behavior of each joint is programmable through interchangeable elements with precomputed stiffness characteristics, enabling users to specify desired robot shapes and deformation profiles in a principled manner.

Beyond mechanical modularity, the platform enables self pose estimation with onboard sensing, avoiding external tracking. We use magnetic sensing with machine learning to infer the configuration of each joint, and keep the learning pipeline modular by training one model on the movements of a single joint that can be reused across different robot joints. This reduces training effort and supports scalable reconfiguration while maintaining reliable pose estimates when trained on sufficiently diverse coil data.

We validated the platform in experiments that culminated in a grasping task that demonstrated both mechanical adaptability and effective self-sensing. These results show the feasibility of combining modular continuum hardware with embedded learning-based proprioception, enabling closed-loop control and greater autonomy in future work.

In summary, the main contributions of this paper are as follows:
\begin{enumerate}
    \item The design of a modular, reconfigurable continuum robotic platform.
    \item A method for programming robot shape through interchangeable joints with stiffness properties that enhance performance for priority tasks while maintaining broad functionality.
    \item A self-contained pose estimation approach using magnetic sensing and modular machine learning models trained at the joint level.
    \item Experimental validation of the proposed system in a set of traversal and grasping tasks.
\end{enumerate}

The segment firmware, data-collection and training code, the pretrained model and the ROS~2 deployment package as well as mechanical design files are released at \url{https://ansue1234.github.io/ModTenV1/}.

\section{RELATED WORKS}
Previous works we took inspiration from which relates to our proposed platform are mainly in three directions, including modular designs for scalable reconfiguration, methods for regulating mechanical stiffness in compliant
robots, and pose and shape estimation.

\subsection{Modular Design}
Current modular designs used in continnum robots includes modular segments for integration into general-purpose manipulators or for locomotion \cite{castledine_design_2019, fang2025hierarchically, cai2025modular}, lightweight tendon-driven backbones built from 3D-printable compliant mechanism units \cite{dewi_lightweight_2024, dewi2026craft-bdd}, and discrete tendon-driven modules using ball-socket joints with superelastic Nitinol rods \cite{kim_planar_2024}. Beyond hinge-based designs, origami-inspired modular deployable structures such as spherical linkage parallel mechanisms have also been proposed to enhance motion accuracy and stability \cite{jia_design_2024}.

\subsection{Stiffness Tuning}
The inherent structural flexibility of continuum robots is advantageous for maneuvering through confined environments, but often leads to reduced axial/longitudinal stiffness and load-bearing capacity. Researchers from \cite{dewi_lightweight_2024, dewi2026craft-bdd} enable pre-programmable stiffness in tendon-driven robots using customized modular compliant units. Other works vary bending stiffness along the body to achieve non-uniform curvature and programmable shapes, often via frictional locking or jamming of proximal/intermediate segments while keeping distal sections flexible \cite{shen_design_2023, yang2025design-0fd, zhang_preprogrammable_2023}. However, these approaches typically rely on rigid or complex structures, while soft-material variants  \cite{wang2024soft-fc5, wang2024spirobslogarithmicspiralshapedrobots} tend to be less modular and reconfigurable.

\subsection{Pose Estimation}
Accurate pose estimation is essential for continuum robots because their hyper-redundant, flexible structures introduce significant uncertainty. Early vision-based methods used supervised learning with stereo vision to map robot configurations into feature spaces and improve robustness to occlusion \cite{reiter_learning_2011}, while later work combined displacement sensors with onboard visual tracking for real-time section-wise tip estimation \cite{fang_design_2023}. However, vision based system require external setup which is not always available. Therefore in-situ pose estimation is essential. Techniques using embedded strain sensors combined with probabilistic approaches have also been explored, such as Gaussian process regression on SE(3) for estimating continuous shape and strain from sparse, noisy measurements \cite{lilge_continuum_2022, teetaert_stochastic_2025}. To avoid the limitations of external sensing, integrated magnetic sensing has emerged as a promising route to proprioception, with prior work showing that learning-based models can recover 3D shape from highly nonlinear magnetic field measurements \cite{baaij_learning_2023, guo2019continuum}, however, requiring detailed kinematic priors.

\section{METHODOLOGY}
\subsection{Modular Design and Fabrication}

\subsubsection{Overview}
\begin{figure*}
\centering
        \includegraphics[width=\textwidth]{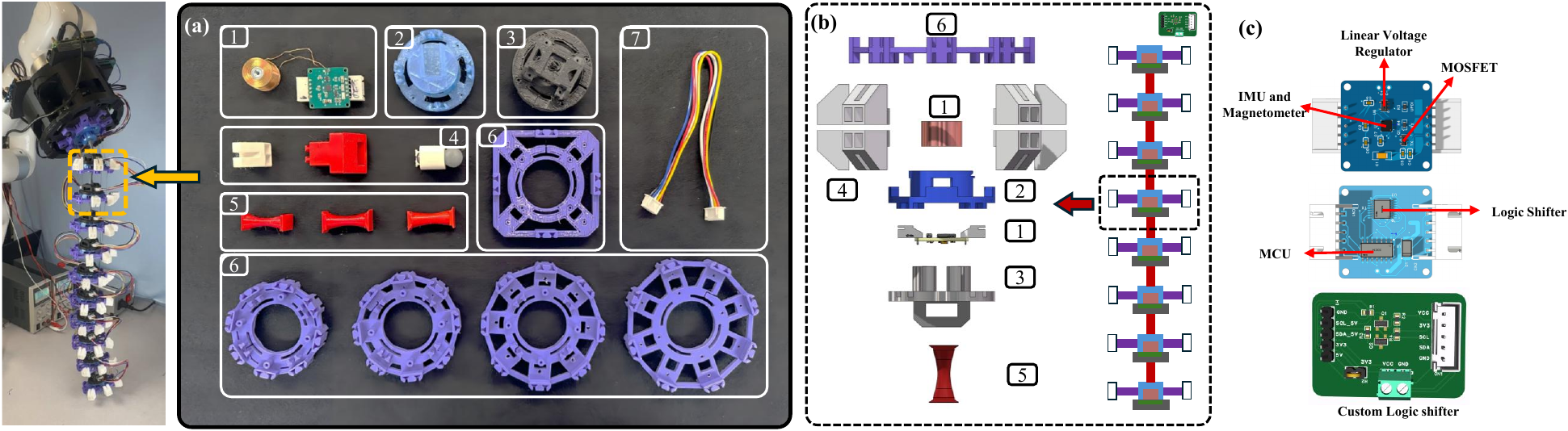}
    \caption{Overview of the plug-and-play components used to assemble the proposed self-sensing modular continuum robotic platform, including the 3-D-printed structural parts and custom electronics. (a) Photograph of the full component set. Numbered parts include (1) sensing module, (2) sensor carrier, (3) cap, (4) stopper insert, (5) flexible joint, (6) limb modules in different sizes, and (7) wiring harness.  (b) Exploded view of a single module and its integration into a multi-segment assembly. (c) Custom sensing boards used in each module. The custom electronics integrate the IMU, magnetometer, voltage regulation, switching, and microcontroller components required for onboard self-sensing. We also make use of a custom logic shifter.}
    \label{fig:structure}
\end{figure*}

Modularity is a central feature of our continuum robot platform. The robot (Fig.~\ref{fig:structure}) comprises identical ``rigid'' core segments, which are 3D printed carbon-fiber reinforced PET, that are connected by compliant 3D printed TPU joints. Each core segment houses the electronics for pose estimation and includes dovetail features for joint attachment, as well as outer slots for interchangeable PLA rings whose diameter and geometry can be customized for specific applications. Rings can also accept task-specific accessories (e.g., high-friction studs for gripping).  This architecture supports an arbitrary number of segments, limited primarily by tendon actuation. In this study, we used 8- and 10-segment robots actuated by servo-driven tendons to demonstrate variable joint stiffness and the resulting sensing and control performance in example grasping tasks.

\subsubsection{Joint Design and Bending Stiffness Characterization}

The modular nature of the design enables the use of joints made from a wide variety of materials and longitudinal and cross-sectional form factors to balance local bending, axial and torsional stiffness.  In this initial study, we used 3D printed TPU joints with a simple ``hourglass"-like shape whose stiffness is dictated largely by its minimum ``neck" diameter.  As shown in Fig. \ref{fig:overall}, joints of varying neck diameters can be used to produce shapes with non-uniform curvature.
The bending stiffness of the joints connecting the modular rigid segments was estimated using finite element studies performed in ANSYS/Mechanical (ANSYS, Inc.).  Mechanical properties of the TPU joints were determined based on the results of uniaxial tension testing using specimens and procedures comparable to ASTM Standard D412 \cite{ASTMD412-2021}. For neck diameters ranging from 3 - 7 mm, we analyzed a single segment of the robot with a three-dimensional, static and geometrically nonlinear model subjected to a 30 degree bend angle on one end and a fixed boundary condition on the other.  The reaction moment (\begin{math}M\end{math}) at the fixed end was determined at every three degrees of bend angle (\begin{math}\theta\end{math}).   The bending/rotational stiffness of the joint (\begin{math}k\end{math}) was determined by fitting a linear relationship between the reaction moment (\begin{math}M\end{math}) and bend angle (\begin{math}\theta\end{math}) according to 
\begin{equation}
    M = k\theta \label{eq:stiff coef}\,
\end{equation}
Future studies will extend this analysis scheme to examine how joints whose behavior is nonlinear due to material or geometric design choices may be leveraged to further ``program" the robot shape for different applications.

\subsubsection{Electronics and Sensor Design}
In each segment, a custom designed printed circuit board (PCB) is embedded to sense the magnetic field of the coil of the nearby segments. Each PCB includes a microcontroller (ATTiny3224), a 9 DoF IMU that includes magnetometer readings (ICM-20948), a linear drop out regulator (TPS72118DBVT), a logic shifter (SN74AXC4T774), as well as a diode and a N Channel MOSFET (NMOS) to control the coil. The coil used in our segments are cylindrical with a height of 12 mm and diameter of 19 mm with 27 AWG wires.

The microcontroller on the PCB acts as a bridge that transfers SPI communication to the actual IMU and I2C communication with a Arduino MEGA to relay the sensor data as well as to activate and deactivate the coil during sensing. Each PCB is daisy chained with JST XH 5 Pin connectors that includes the power lines for the coil and onboard components as well as I2C communication lines between each component and the Arduino MEGA that interfaces with the Raspberry Pi for pose inference. 

Because the IMU operates on 1.8V logic, while the rest of the circuit, including the microcontroller, operates on 3.3V, a logic shifter is needed to convert the voltages during communication between these two devices. Furthermore, the Arduino MEGA operates on 5V logic, hence another custom logic shifter board is designed using NMOS (BSS138) to interface the Arduino MEGA with all of the microcontrollers. A schematic of both boards are in Fig. \ref{fig:structure}(c).

\subsection{Pose Estimation}
In contrast to previous continuum-robot self-pose sensing based on distributed strain gauges or embedded cameras/accelerometers \cite{fang_design_2023,lilge_continuum_2022,stella_soft_2024}, we use magnetic sensing by embedding magnetometers and field sources within the robot, similar in spirit to \cite{baaij_learning_2023}. Unlike \cite{baaij_learning_2023}, which uses permanent magnets, we employ electromagnetic coils that can be selectively activated for sensing and deactivated to reduce inter-segment interference.

\subsubsection{Working Principle}

\begin{figure*}
    \centering
    \includegraphics[width=0.85\textwidth]{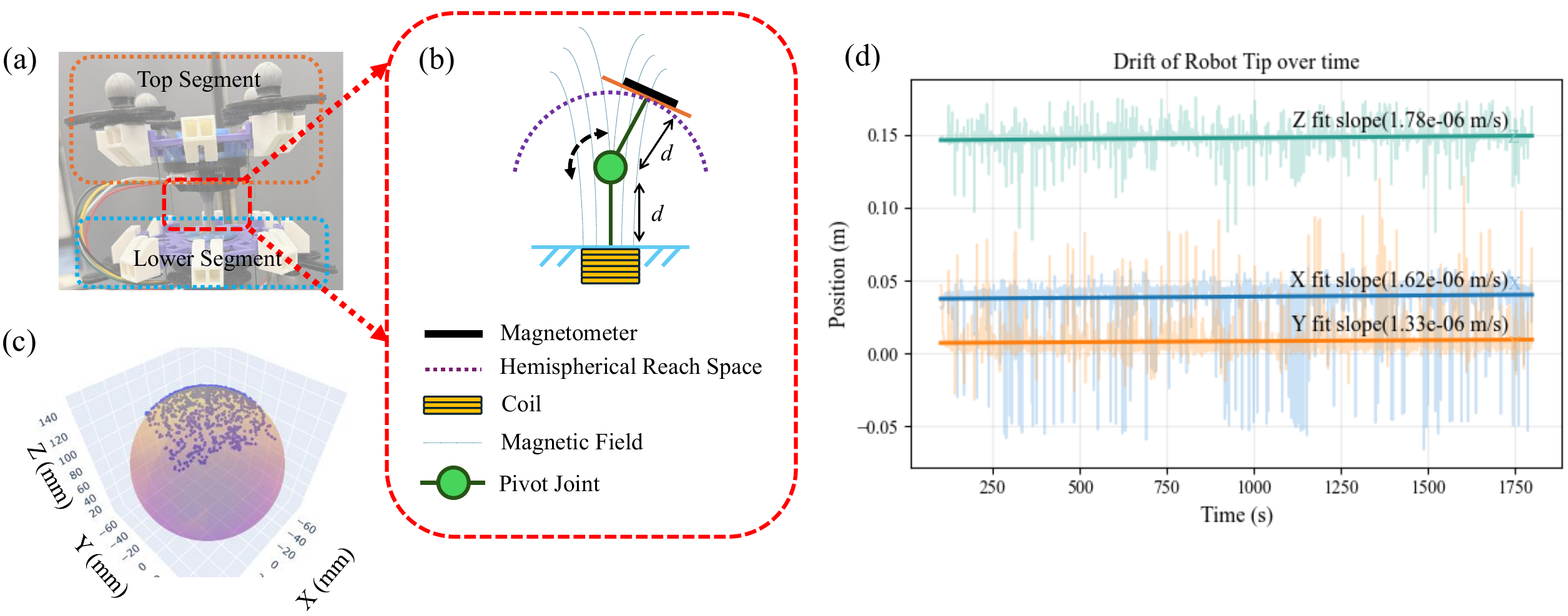}
    \caption{(a) Example of a pair of segments and joint within the robot. In this view, the magnetometer is on the top segment and the coil is on the bottom segment. (b) Schematic of the joint motion. We consider the soft joint as a rigid hinge with a fixed rotation axis at the joint center and minimal compression (constant $d$).  The tip position is inferred from magnetometer readings as it moves through the coil’s field. (c) Motion-capture data of tip positions from one run (500 waypoints), showing that tip positions lie on a a hemisphere with 0.523 mm RMSE. (d) We lock the robot in a stationary pose for 26 minutes and observed negligible drift. 
}
    \label{fig:working_principle}
\end{figure*}

Our pose estimation maps magnetometer measurements in an actively generated magnetic field to relative position via machine learning. As illustrated in Fig.~\ref{fig:working_principle}(b), when adjacent segments bend about a compliant joint, the magnetometer in one segment measures the field produced by the coil embedded in its neighbor. We assume: (i) joints maintain a fixed arc length, with bending as the dominant deformation mode and (ii) joints pivot about their geometric center during bending. Together, these assumptions constrain inter-segment motion to a dome-like (hemispherical) manifold. Our joint design is intended to enforce these conditions, and Fig.~\ref{fig:working_principle} (c) empirically supports their validity. Under these constraints and assumptions, we adopt a data-driven approach to determine position based on magnetic field measurements whereby we collect magnetic field data over the reachable relative workspace via randomized ``wobble'' motions and train a neural network to learn the mapping from field measurements to position.

Because all segments use identical manufacturer-matched electromagnetic coils, we train a single joint-level network over one joint's reachable workspace and reuse it across all joints, yielding a modular sensing pipeline. Since each pair of adjacent segments is treated independently, the method scales to robots with an arbitrary number of segments.

\subsubsection{Data Collection}
Training data for the magnetic to position mapping were collected using a two-segment setup. In each run, the lower segment was rigidly mounted to a fixed base to provide a stationary reference frame, while the upper segment was connected via a compliant joint and allowed to rotate freely under tendon actuation. A diagram of the setup is shown in Fig.\ref{fig:working_principle} (a). 

During data collection, we execute the following sampling sequence. The procedure begins by moving all four motors $\mathcal{I}=\{M_1,M_2,M_3,M_4\}$ to their center positions $\{c_i\}$. For each waypoint $k$, two values $(x_k,y_k)$ are sampled uniformly at random from the square $[-1,1]\times[-1,1]$. These values are scaled by an amplitude $A$ and used to generate target motor commands as symmetric offsets about the motor centers; the resulting commands are clipped to the valid range $[0,4095]$. The motors are then moved to these target positions.

Once the robot reaches the waypoint, the coil is kept \emph{off} and the ambient magnetic bias $\mathbf{b}_k$ is estimated by averaging 10 sensor samples, with a short delay $\Delta t$ between samples. Next, the coil is turned \emph{on} and a short burst of sensor readings $\{\mathbf{s}_{k,j}\}$ (10 samples total) is recorded, again with delay $\Delta t$ between samples. Finally, the system logs the waypoint index $k$, the sampled inputs $(x_k,y_k)$, the motor state (e.g., $q$), the bias $\mathbf{b}_k$, and the full coil-on burst $\mathbf{s}_{k,*}$, and then proceeds to the next waypoint. After every 500 waypoints, the system pauses for a 10-minute cooldown period.

Although the coils are manufacturer-matched, small coil-to-coil variations remain. To promote generalization, we cyclically permuted segment roles across runs. After each run, the previously actuated upper segment became the fixed lower segment, and a new segment was introduced as the actuated upper segment. Each run samples 500 3D positions, and at each position we record 10 magnetometer and accelerometer measurements. In total, we collected data at $10^4$ positions and approximately $10^5$ magnetometer samples across all 10 segments.

To mitigate errors from tendon slack, we do not use the commanded motor positions as ground truth. Instead, we use the accelerometer measurements recorded concurrently with the magnetometer data to recover the corresponding 3D ground-truth pose for each waypoint. During each measurement burst the joint is held stationary, so the accelerometer measures gravity alone; since gravity is perpendicular to the ground plane, we use this constraint to infer the ground truth joint pose.

Since we assume that the movement space of a joint is part of a hemisphere, we find an azimuth angle $\theta$ and a colatitude angle $\lambda$ to represent the position in spherical coordinate that corresponds to each magnetometer reading. Furthermore, our assumption implies a fixed radius and that there is zero yaw angle in the radial axis in the magnetometer or accelerometer body frame due to our tendon driven activation which solely induces translational motions, we adopt the following formulation to find the angles.

For each sample, let the measured acceleration in the sensor/body frame be
\begin{equation}
    \mathbf a_b =
        \begin{bmatrix}
        a_{x,b}, 
        a_{y,b}, 
        a_{z,b}
        \end{bmatrix}^T ,
\end{equation}
We use a world-aligned right-handed frame with axes $(X,Y,Z)=(\text{West},\text{North},\text{Out})$. The sensor axes are mounted as $(x,y,z)=(\text{West},\text{South},\text{Inside})$.
Define $\mathbf a$ as acceleration in the world frame
\begin{equation}
    \mathbf a=
        \begin{bmatrix}
        a_x\\
        a_y\\
        a_z
        \end{bmatrix}
        =
        \begin{bmatrix}
        1&0&0\\
        0&-1&0\\
        0&0&-1
        \end{bmatrix}
        \mathbf a_b ,
\end{equation}
Since the angles depend only on direction, hence we normalize the gravity vector that is detected from the accelerometer.
\begin{equation}
    \hat{\mathbf r}=\mathbf a/\|\mathbf a\|_2 =
        \begin{bmatrix}
        r_x, 
        r_y,
        r_z
        \end{bmatrix}^T ,
\end{equation}
Therefore the colatitude $\lambda$ (0 at top, measured from $+Z$):
\begin{equation}
\lambda=\arccos\!\Big(\mathrm{clip}\big(r_z,-1,1\big)\Big).
\label{eq:colatitude}
\end{equation}
The azimuth is obtained from the roll and pitch of the gravity vector, $\phi=\operatorname{atan2}(r_y, r_z)$ and $\psi=\operatorname{atan2}\big(-r_x,\sqrt{r_y^2+r_z^2}\,\big)$, whose radial direction $(\sin\psi\cos\phi,\;\sin\phi,\;\cos\psi\cos\phi)$ points from the virtual pivot toward the tilted segment. Taking its horizontal components gives the azimuth $\theta$ with $\theta=0$ along North ($+Y$) and increasing toward West ($+X$):
\begin{equation}
\theta=\operatorname{atan2}\!\big(-\,r_x\,r_z,\;r_y\big),
\label{eq:azimuth}
\end{equation}
where $\theta \in [-\pi, \pi)$ and $r_z\ge 0$ on the hemisphere.\footnote{Erratum: an earlier version of this manuscript stated $\theta=\operatorname{atan2}(r_x,\,r_y)$. That expression is mirrored about the North axis with respect to what the released code, the training labels and the released model actually use, which is Eq.~(\ref{eq:azimuth}); for small tilts the two differ by the sign of $\theta$. Eq.~(\ref{eq:azimuth}) is the correct one, and the colatitude in Eq.~(\ref{eq:colatitude}) is unaffected.} This formulation helps the model to be more general, able to adapt to rigid segments of arbitrary lengths while keeping the joint positions accurate.

\subsubsection{Setup and Training}
As there is no temporal dependence between the magnetic field and position, we decided to use an MLP (Multi-Layer Perceptron) architecture with a drop out layer to act as the model for the relationship between magnetometer reading and position. 

Although we represent the position data as a coordinate of the azimuth angle $\theta$ and the colatitude angle $\lambda$, due to $\pi$ wrapping of the azimuth angle, there are discrete jumps which are difficult for the neural network to handle; therefore, we transform the two angles into a 3D direction vector, which points from the center virtual pivot of the joint to a point on the hemispherical reach space instead as the ground truth for the neural network to predict. The transformation is as follows:
\begin{equation}
    \mathbf d(\theta,\lambda)=
    \begin{bmatrix}
    \sin\lambda \,\cos\theta\\[4pt]
    \sin\lambda \,\sin\theta\\[4pt]
    \cos\lambda
    \end{bmatrix} ,
\end{equation}
where $\mathbf d$ is the direction vector for prediction from a magnetometer reading. In addition, to help with generalization, we used a normalize each magnetometer reading and feed the normalized reading as the input to predict the direction vector. Therefore, our input for the neural network has a dimension of 3, the normalized magnetometer, and the output of our network has a dimension of 3 as well, the normalized direction vector.

For training the neural network, we used a cosine loss, defined as the following:
\begin{equation}
\ell_{\cos} = 1 - \mathrm{clip}\!\left(\hat{\mathbf m}^\top \hat{\mathbf d},\,-1,\,1\right) 
\end{equation}
where $\hat{\mathbf m}_i$ and $\hat{\mathbf d}_i$ represent the normalized predicted direction vector from the magnetometer readings and the ground truth direction vector, respectively.

We randomized the data collected across all coils and used 85\% of the data for training and 15\% as a holdout set. We optimized and trained the network using AdamW. Using Bayesian optimization in Optuna (10-hour search, up to 500 epochs per trial with early stopping and pruning, single Nvidia RTX~5090), we found the best hyperparameters to be: network dimensions $[1078, 222, 212, 346]$ with GELU activation, dropout $= 3.89 \times 10^{-4}$, no BatchNorm, learning rate $= 5.62 \times 10^{-4}$ with decay $= 2.31 \times 10^{-3}$, and batch size $=256$.

\subsubsection{Inference}
In real life inference of the joint position, we deploy our trained model in ONNX format on a Raspberry Pi 5 8GB Variant (RPi) running a Docker container with ROS2 Humble. An Arduino MEGA acts as a communication bridge between the magnetometers and the RPi. 

To minimize magnetic interference between the joints in our full length robot, we turn only one coil on and collect the magnetometer data of segment that neighbors the the coil at any given time. For each joint, we spend 20 ms for the magnetic field in the activated coil to settle down. The process is repeated across all the segments along the length of the robot in a sequential fashion. In detail, we used the following sampling scheme to acquire the magnetometer data needed for pose estimation. Including the coil settling time and overhead in communication, it takes approximately 80 ms to sample a single joint. 

The inference sampling procedure runs continuously to reduce inter-segment magnetic interference by exciting only one coil at a time. At the start of each cycle, all coils on the $N$ bridges (with I$^2$C addresses $\{a_i\}_{i=0}^{N-1}$) are turned \emph{off}. The bridge to sample is selected sequentially as
\begin{equation}
    j = (\texttt{cycle} \bmod N) + 1,
\end{equation}
so that $j$ cycles through $1,\dots,N$.

With all coils off, the system waits an ambient settling time $t_{\text{amb}}$, then reads the IMU from bridge $j$. From the returned packet $\mathbf{s}_j$ (accelerometer/gyroscope/magnetometer), the magnetometer vector is extracted and used as the ambient magnetic bias $\mathbf{b}_j$ for that bridge. After a short delay $t_{\Delta}$, the coil associated with bridge $(j-1)$ is turned \emph{on}, the system waits a coil-field settling time $t_{\text{coil}}$, and the IMU on bridge $j$ is read again.

From this second readout, the raw magnetometer vector $\mathbf{m}^{\text{raw}}_j$ is extracted and debiased via
\begin{equation}
    \mathbf{m}^{\text{deb}}_j = \mathbf{m}^{\text{raw}}_j - \mathbf{b}_j.
\end{equation}
The resulting debiased measurement is stored in the sequential array as $\texttt{seq}[j-1] \leftarrow \mathbf{m}^{\text{deb}}_j$. When $j=N$, the full array $\texttt{seq}$ is transmitted for pose inference. Finally, the coil $(j-1)$ is turned \emph{off}, $\texttt{cycle}$ is incremented, and the procedure repeats indefinitely.

\begin{figure*}
    \centering
    \includegraphics[width=0.85 \textwidth]{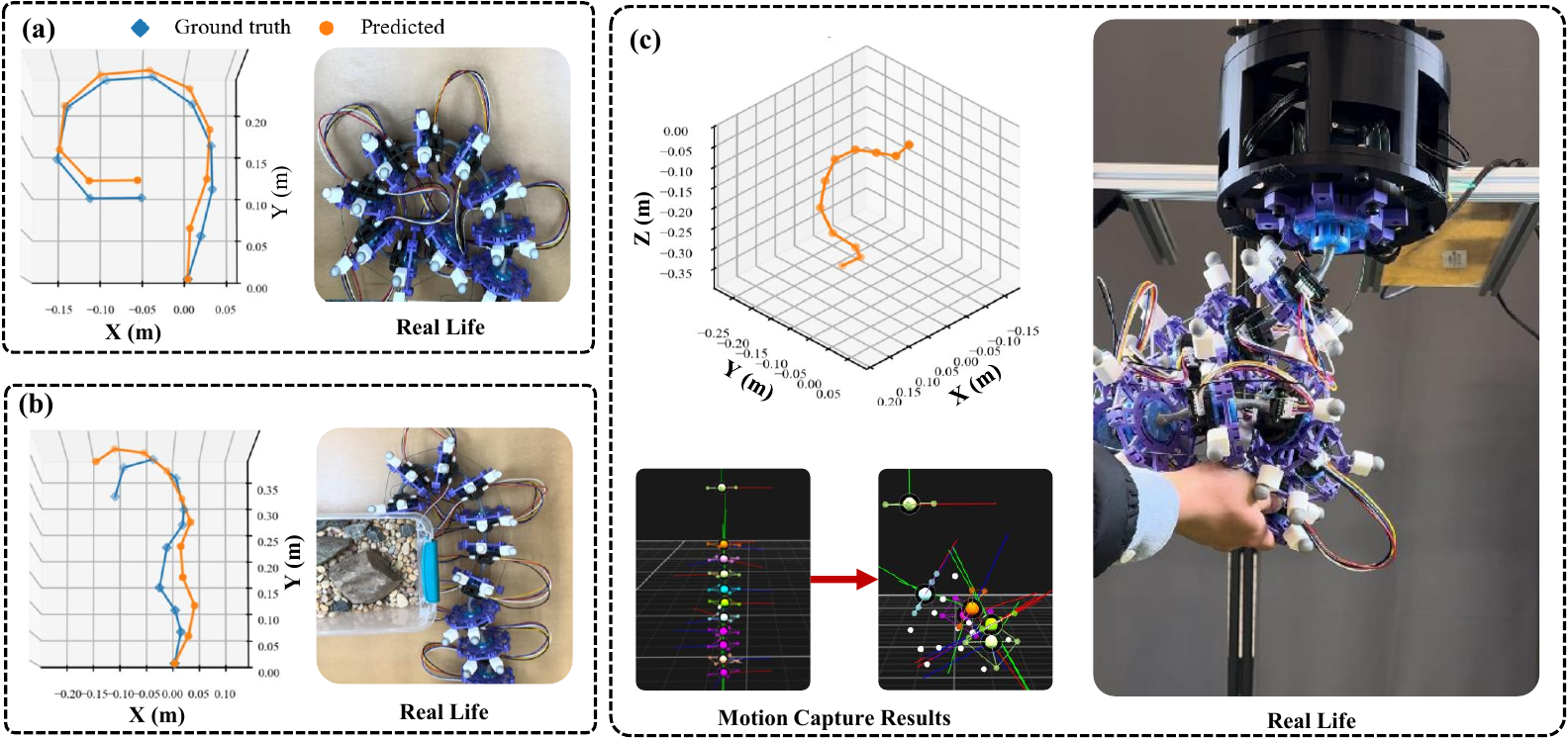}
    \caption{(a) Pose estimation results of Scenario 1 where the robot is allowed to move freely without restriction. The computed RMSE of the prediction to ground truth is 0.025483 m. (b)Pose estimation results of Scenario 2 where the robot movement is blocked by a box of rocks. The computed RMSE of the prediction to ground truth is 0.0395 m. (c) Pose estimation results of Scenario 3 where the robot position is manipulated by hand. In this scenario the motion capture failed in tracking the reflective markers hence no ground truth is available; however, our pose estimation still outputs a relatively accurate result under such large deformation.}
    \label{fig:pose_estimate_results}
\end{figure*}

As inference is only done when $\texttt{seq}$ is received, i.e. all segment's magnetometer data have been read, the sampling rate of our pose estimation is around 1.2 Hz due to communication overheads, magnetometer sampling rates and the physical magnetic field settling time of the coils when activated $t_{\text{coil}}$ and deactivated $t_{\text{amb}}$, which is around 20 milliseconds each.

\subsubsection{Motion Capture System}

We used an Optitrack Motive motion capture system to capture the position of the robot during deformation as the ground truth for pose estimation studies.

\section{RESULTS}
\subsection{Pose Estimation}
We used three scenarios to evaluate and validate our pose estimation scheme: (1) unobstructed planar motion, (2) impeded planar motion, and (3) manual 3D manipulation to arbitrary shapes.

\begin{figure}[!t]
\centering
        \includegraphics[width=\columnwidth]{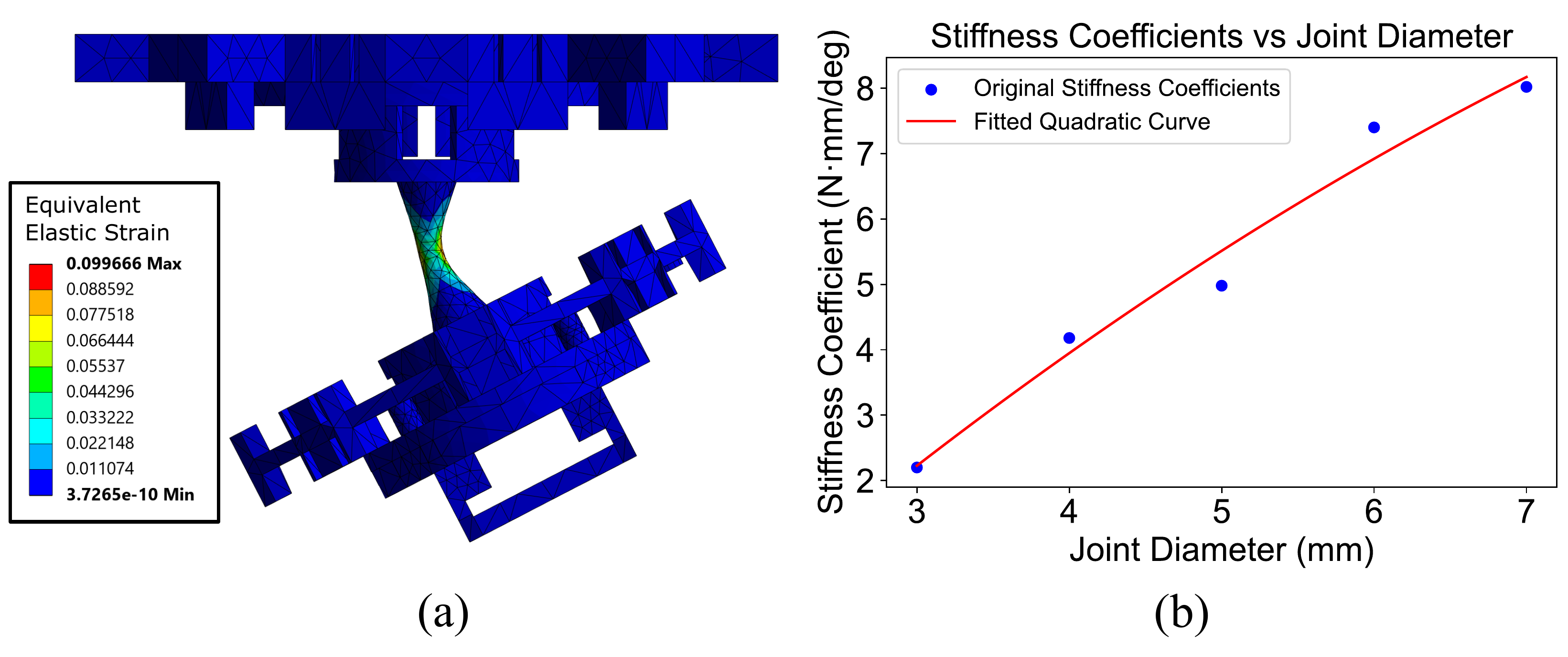}
    \caption{(a) Exemplar strain contours of finite element studies; (b) Bending stiffness of joints with varying minimum diameters}
    \label{fig:fea}
\end{figure}

\begin{figure}[!t]
\centering
        \includegraphics[totalheight=20cm]{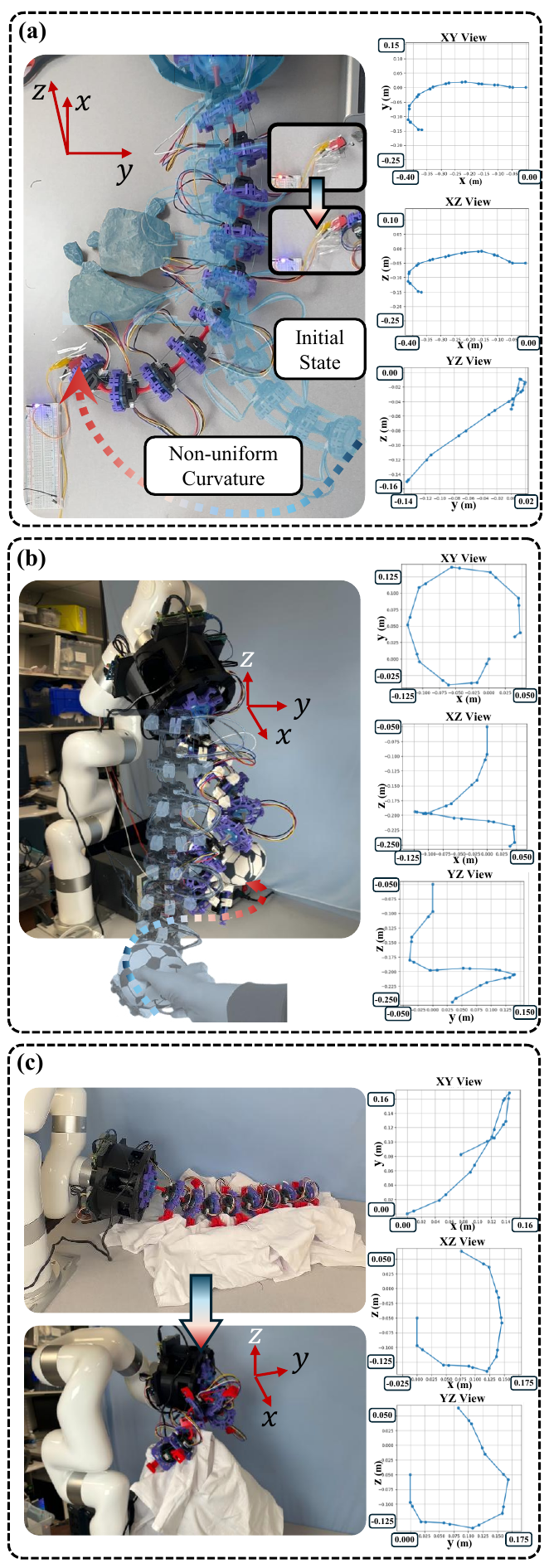}
    \caption{Demonstrations of our modular continuum robotic platform with corresponding pose estimates. (a) Variable stiffness joints are used to traverse an obstacle and contact a target; (b) ``Stoppers" with softer, higher friction surfaces are added to enable grasping of a soft, spherical object; \textbf{Bottom}: Alternative limb/rings are used to grasp a flexible sheet/cloth.}
    \label{fig:demo}
\end{figure}

\subsubsection{Unobstructed Planar Motion}
In Scenario~1, the robot is laid down on the ground and free to move on the ground without obstructions. We activated one tendon, and observed the behavior of the pose estimation scheme as the robot curl up. This is the simplest test that probes the ability of our pose estimation without external influences. The setup and result can be seen in Fig.\ref{fig:pose_estimate_results}(a).

\subsubsection{Impeded Planar Motion}
In Scenario~2, the robot is placed on the ground with its motion impeded by a box of rocks. We actuate a single tendon and evaluate the pose estimator as the robot deforms under obstacle contact forces. This tests robustness to unmodeled disturbances in a nominally planar setting. The setup and results are shown in Fig.~\ref{fig:pose_estimate_results}(b).

\subsubsection{Manual 3D Manipulation}
In Scenario~3, the robot is suspended vertically and manually manipulated by hand to bend and curl in 3D. Self-occlusion during this motion prevents the motion-capture system from reliably tracking markers, yielding no usable ground truth (Fig.~\ref{fig:pose_estimate_results}(c)). Nevertheless, our trained sensing scheme still produces pose estimates that appear qualitatively accurate by visual inspection.

\subsection{Joint Bending Stiffness}
An exemplar deformed shape and strain contour plot of a rotated joint segment is shown in Fig.~\ref{fig:fea} along with the joint bending stiffness for joints ranging from 3 - 7 mm neck diameter.  A quadratic fit of the stiffness trend is shown to illustrate how these and similar analyses could be used in a design setting to determine the exact neck diameter needed to give a specific bending stiffness required for a given application.

\subsection{Application Demonstrations}

To evaluate the practical versatility of the proposed design, we configured the robot with different combinations of limb, joint, and stopper modules and tested it in three representative manipulation tasks. These experiments were intended to demonstrate the reconfigurability of the platform and its ability to adapt to different task requirements. Representative results are shown in Fig.~\ref{fig:demo}.

First, as shown in Fig.~\ref{fig:demo}(a), we combined limbs with different effective lengths and flexible joints with different bending stiffness to promote a non-uniform curvature under tendon actuation.  The choice of effective lengths and stiffness enabled the robot to bend around an obstacle and reach a target position for a switch. 

Second, we performed a spherical grasping task using TPU stopper modules coated with Eco-Flex 00-30 (Smooth-On, Inc), as shown in Fig.~\ref{fig:demo}(b). This configuration provided both compliant contact and increased surface friction, allowing the robot to stably envelope and grasp a sphere with a diameter of approximately 12~cm. The result demonstrates how simply replacing contact modules can substantially improve grasping performance for smooth curved objects.

Third, we used square limb modules together with red stopper inserts designed to provide larger gripping force for cloth grasping, as shown in Fig.~\ref{fig:demo}(c). Compared with the previous configuration, this setup generated stronger local contact forces and was better suited for handling deformable objects. This further demonstrates that the same platform can be rapidly reconfigured to accommodate different object geometries and contact conditions.

In all three tasks, we used the proposed pose estimation method to predict the robot shape during operation. The results show that the method can effectively estimate both the relative positions between neighboring segments and the overall body configuration while maintaining good performance during real contact-rich interactions. In general, these demonstrations validate the effectiveness and generality of the proposed self-sensing modular continuum robotic platform in diverse manipulation tasks.

\section{DISCUSSION}
\subsection{Robot Design}

\subsubsection{Segment and Joint Modularity}

This design eliminates the need for threaded fasteners and allows each segment to be assembled and disassembled without tools. As a result, structural parts, stiffness elements, and task-specific attachments can be replaced at the module level with minimal disturbance to the sensing hardware. During these initial studies, this facilitated the reconfiguration effort, simplified maintenance, and supported rapid hardware iteration across different experimental setups.  However, we did observe areas for improvement. That is, dovetail connections sometimes "popped" out under high loads. Additionally, the cumulative weight of the system limited its functionality.  These issues can be addressed in future versions of the platform.

\subsubsection{Joint Stiffness Variation and Characterization}

In this initial study, variations on a single relatively simple joint geometry were used to demonstrate how joint bending stiffness could be varied along the length of the robot in order to promote shapes with non-uniform curvature. Although this is in itself not unique, the same modular design scheme, materials, and analysis pipeline can be leveraged to produce more sophisticated and complex joint designs for more intricate shapes and and tasks.  Doing so will likely require accounting for the non-linear deformation of the joint.   As is, the current study demonstrates that the modular design and the variable joint stiffness approach are compatible with our onboard pose estimation scheme.

\subsection{Pose Estimation}

Our experiments show that the proposed magnetic self-pose estimator captures robot shape across various conditions and, unlike other magnetic approaches \cite{wang2024spirobslogarithmicspiralshapedrobots,baaij_learning_2023}, requires no kinematic prior. Scenario~1 validates the method in an unobstructed setting. In Scenario~2 (planar obstruction), the estimator remains accurate under disturbances that cause complex unknown kinematics such as forced deformations of contact forces. Scenario~3 further removes any kinematic observability by not utilizing tendon actuation and instead moving the robot by hand. Despite motion-capture failure, the system still produces qualitatively correct pose estimates, highlighting the value of in-situ sensing relative to vision-based methods. Quantitatively, our RMSE is comparable to accelerometer-based self-pose sensing reported in \cite{stella_soft_2024}, indicating that magnetic sensing is a viable alternative for tendon-driven modular robots.

\subsection{Limitations and Future Work}

Although our pose estimation achieves an accuracy comparable to state-of-the-art methods, several limitations remain. First, the coil for magnetic field generation is bulky (19\,mm $\times$ 12\,mm). Therefore, replacing discrete coils with PCB-embedded copper-trace coils could reduce size, weight and improve integration. Second, while the interference-mitigation sampling scheme improves signal quality, it makes the global sampling rate scale poorly with robot length. For our 10-linkage system, the maximum stable rate is $\sim$1.2\,Hz, which is insufficient for tasks that requires fast movements such as pick and place in a factory setting. Parallelized or faster acquisition schemes are therefore an important direction for future work.

\section{Conclusion}

In this work, we present the design of a modular, tendon driven continuum robot with variable stiffness joints and embedded self pose sensing capabilities. We demonstrate how a mechanical analysis can be used to tailor a particular configuration of joints to produce a desired actuated shape, and the combined modularity and variable stiffness functionality was confirmed to be compatible with a novel self pose estimation scheme utilizing magnetic sensing and machine learning.  This unique combination of modularity, variable stiffness, and in-situ pose estimation establishes our design as a promising platform for future research on the capabilities of continuum robot systems.


\addtolength{\textheight}{-12cm}   




\section*{ACKNOWLEDGMENT}
This work used large language models (LLMs) such as ChatGPT, Gemini, and Claude to assist in proofreading and revising the manuscript text.  In addition, LLMs were also used to assist in the creation of code segments/scripts used for experimental data analysis and figure generation.


\bibliographystyle{IEEEtran}
\bibliography{reference}

\end{document}